\documentclass{article} 
\usepackage{nips14submit_e,times}
\usepackage{hyperref}
\usepackage{url}
\usepackage{natbib}
\usepackage{amsmath}
\usepackage{amsthm}
\usepackage{amssymb}
\usepackage{graphicx}
\usepackage{xspace}
\usepackage{tabularx}
\usepackage{enumitem}
\usepackage{ownstyles}

\usepackage{xr} 

\newcommand{\net}[1]{\ensuremath{\mathsf{#1}\xspace}}

\newcommand{\dA}{\net{A}\xspace}
\newcommand{\dB}{\net{B}\xspace}
\newcommand{\capemph}[1]{\emph{#1}}

\usepackage{fancyhdr}
\fancypagestyle{firstpagestyle}
{
  \fancyhead[C]{
    \hspace*{-.15\textwidth}\raisebox{-2em}[0pt][0pt]{\parbox{1.3\textwidth}{\sffamily Preprint release. Full citation: Yosinski J, Clune J, Bengio Y, and Lipson H. \textit{How transferable are features in deep neural networks?} In Advances in Neural Information Processing Systems 27 (NIPS '14), NIPS Foundation, 2014.}}
  }
}

\title{How transferable are features in deep neural networks?}

\iftrue
\author{
Jason Yosinski,$^1$ \ \ Jeff Clune,$^2$ \ \ Yoshua Bengio,$^3$ \ and \ Hod Lipson$^4$ \\
$^1$ Dept. Computer Science, \ \ Cornell University \\
$^2$ Dept. Computer Science, \ \ University of Wyoming \\
$^3$ Dept. Computer Science \& Operations Research, \ \ University of Montreal \\
$^4$ Dept. Mechanical \& Aerospace Engineering, \ \ Cornell University \\
}
\fi

\nipsfinalcopy 

\begin{document}

\maketitle

\thispagestyle{firstpagestyle}

\begin{abstract}

Many deep neural networks trained on natural images exhibit a curious
phenomenon in common: on the first layer they learn features similar
to Gabor filters and color blobs.  Such first-layer features appear
not to be \emph{specific} to a particular dataset or task, but
\emph{general} in that they are applicable to many datasets and tasks.
Features must eventually transition from general to specific by the
last layer of the network, but this transition has not been studied
extensively.  In this paper we experimentally quantify the generality
versus specificity of neurons in each layer of a deep convolutional
neural network and report a few surprising results. Transferability is
negatively affected by two distinct issues: (1) the specialization of
higher layer neurons to their original task at the expense of
performance on the target task, which was expected, and (2)
optimization difficulties related to splitting networks between
co-adapted neurons, which was not expected.  In an example network
trained on ImageNet, we demonstrate that either of these two issues
may dominate, depending on whether features are transferred from the
bottom, middle, or top of the network.  We also document that the
transferability of features decreases as the distance between the base
task and target task increases, but that transferring features even
from distant tasks can be better than using random features. A final
surprising result is that initializing a network with transferred
features from almost any number of layers can produce a boost to
generalization that lingers even after fine-tuning to the target
dataset.

\end{abstract}


\section{Introduction}

Modern deep neural networks exhibit a curious phenomenon: when trained on images, they all tend to learn first-layer features that resemble either Gabor filters or color blobs.
The appearance of these filters is so common that obtaining anything else on a natural image dataset causes suspicion of poorly chosen hyperparameters or a software bug.
This phenomenon occurs not only for different datasets, but even with very different training objectives, including supervised image classification \citep{Krizhevsky-2012}, unsupervised density learning \citep{HonglakL2009}, and unsupervised learning of sparse representations \citep{Le-2011-ICA}.

Because finding these standard features on the first layer seems to occur regardless of the exact cost function and natural image dataset, we call these first-layer features \emph{general}. On the other hand, we know that the features computed by the last layer of a trained network must depend greatly on the chosen dataset and task. For example, in a network with an N-dimensional softmax output layer that has been successfully trained toward a supervised classification objective, each output unit will be specific to a particular class. We thus call the last-layer features \emph{specific}. These are intuitive notions of \emph{general} and \emph{specific} for which we will provide more rigorous definitions below.
If first-layer features are general and last-layer features are specific, then there must be a transition from general to specific somewhere in the network. This observation raises a few questions:

\begin{itemize}[leftmargin=2em]
	\item Can we quantify the degree to which a particular layer is general or specific?
	\item Does the transition occur suddenly at a single layer, or is it spread out over several layers?
	\item Where does this transition take place: near the first, middle, or last layer of the network?
\end{itemize}

We are interested in the answers to these questions because, to the extent that features within a network are general, we will be able to use them for 
\emph{transfer learning} \citep{caruana95,Bengio+al-AI-2011-small,UTLC+DL+tutorial-2011-small}.
In transfer learning, we first train
 a \emph{base} network on a base dataset and task, and then we repurpose the learned features, or \emph{transfer} them, to a second \emph{target} network to be trained on a target dataset and task. This process will tend to work if the features are general, meaning suitable to both base and target tasks, instead of specific to the base task.

When the target dataset is significantly smaller than the base dataset, transfer learning can be a powerful tool to enable training a large target network without overfitting; Recent studies have taken advantage of this fact to obtain
state-of-the-art results when transferring from higher layers \citep{donahue+jia-2013-arxiv,Zeiler+et+al-arxiv2013b,sermanet2013overfeat:-integrated-recognition}, collectively suggesting that these layers
of neural networks do indeed compute features that are fairly general. These results further emphasize the importance of studying
the exact nature and extent of this generality.

The usual transfer learning approach is to 
train a base network and then copy its first $n$ layers to the first $n$ layers of a target network. The remaining layers of the target network are then 
randomly initialized and trained toward the target task. One can choose to backpropagate the errors from the new task into the base (copied) features to \emph{fine-tune} them to the new task, or the transferred feature layers can be left \emph{frozen}, meaning that they do not change during training on the new task. The choice of whether or not to fine-tune the first $n$ layers of the target network depends on the size of the target dataset and the number of parameters in the first $n$ layers. If the target dataset is small and the number of parameters is large, fine-tuning may result in overfitting, so the features are often left frozen. On the other hand, if the target dataset is large or the number of parameters is small, so that overfitting is not a problem, then the base features can be fine-tuned to the new task to improve performance. Of course, if the target dataset is very large, there would be little need to transfer because the lower level filters could just be learned from scratch on the target dataset. We compare results from each of these two techniques --- fine-tuned features or frozen features --- in the following sections.

In this paper we make several contributions:

\begin{enumerate}[leftmargin=2em]

\item We define a way to quantify the degree to which a particular layer is general or specific, namely, how well features at that layer transfer from one task to another (\secref{definition}). We then train pairs of convolutional neural networks on the ImageNet dataset and characterize the layer-by-layer transition from general to specific (\secref{experiments}), which yields the following four results. 

\item We experimentally show two separate issues that cause performance degradation when using transferred features without fine-tuning: (i) the specificity of the features themselves, and (ii) optimization difficulties due to splitting the base network between co-adapted neurons on neighboring layers. We show how each of these two effects can dominate at different layers of the network. (\secref{experiments_ab})

\item We quantify how the performance benefits of transferring features decreases the more dissimilar the base
task and target task are. (\secref{experiments_nm})

\item On the relatively large ImageNet dataset, we find lower performance than has been previously reported for smaller datasets \citep{Jarrett-ICCV2009} when using features computed from random lower-layer weights vs. trained weights. We compare random weights to transferred weights---both frozen and fine-tuned---and find the transferred weights perform better. (\secref{experiments_random})

\item Finally, we find that initializing a network with transferred features from almost any number of layers can produce a boost to generalization performance after fine-tuning to a new dataset. This is particularly surprising because the effect of having seen the first dataset persists even after extensive fine-tuning.
(\secref{experiments_ab})

\end{enumerate}


\section{Generality vs. Specificity Measured as Transfer Performance}
\seclabel{definition}

We have noted the curious tendency of Gabor filters and color blobs to show up in the first layer of neural networks trained on natural images.
In this study, we define the degree of generality of a set of features learned on task \dA as the extent to which the features can 
be used for another task \dB.
It is important to note that this definition depends on the similarity between \dA and \dB. We create pairs of classification tasks \dA and \dB by constructing pairs of non-overlapping subsets of the ImageNet dataset.\footnote{The ImageNet dataset, as released in the Large Scale Visual Recognition Challenge 2012 (ILSVRC2012) \citep{imagenet_cvpr09} contains 1,281,167 labeled training images and 50,000 test images, with each image labeled with one of 1000 classes.}
These subsets can be chosen to be similar to or different from each other.

To create tasks \dA and \dB, we randomly split the 1000 ImageNet classes into two groups each containing 500 classes and approximately half of the data, or about 645,000 examples each.
We train one eight-layer convolutional network on \dA and another on \dB. These networks, which we call \net{baseA} and \net{baseB}, are shown in the top two rows of \figref{transfer}.
We then choose a layer $n$ from $\{1, 2, \ldots, 7\}$ and train several new networks. In the following explanation and in \figref{transfer}, we use layer $n=3$ as the example layer chosen.
First, we define and train the following two networks:

\begin{itemize}[leftmargin=2em]
\item A \emph{selffer} network \net{B3B}:  the first $3$ layers are copied from \net{baseB} and frozen. The five higher layers (4--8) are initialized randomly and trained on dataset \dB. This network is a control for the next transfer network. (\figref{transfer}, row 3)

\item A \emph{transfer} network \net{A3B}: the first $3$ layers are copied from \net{baseA} and frozen. The five higher layers (4--8) are initialized randomly and trained toward dataset \dB. Intuitively, here we copy the first $3$ layers from a network trained on dataset \dA and then learn higher layer features on top of them to classify a new target dataset \dB. If \net{A3B} performs as well as \net{baseB}, there is evidence that the third-layer features are general, at least with respect to \dB. If performance suffers, there is evidence that the third-layer features are specific to \dA. (\figref{transfer}, row 4)
\end{itemize}

We repeated this process for all $n$ in $\{1, 2, \ldots, 7\}$\footnote{Note that $n=8$ doesn't make sense in either case: \net{B8B} is just \net{baseB},
and \net{A8B} would not work because it is never trained on \dB.} and in both directions (i.e. \net{AnB} and \net{BnA}).
In the above two networks, the transferred layers are \emph{frozen}. We also create versions of the above two networks where the transferred layers are \emph{fine-tuned}:

\begin{itemize}[leftmargin=2em]
\item A \emph{selffer} network \net{B3B^+}: just like \net{B3B}, but where all layers learn.
\item A \emph{transfer} network \net{A3B^+}: just like \net{A3B}, but where all layers learn.
\end{itemize}

\begin{figure}[t]
\begin{center}
\includegraphics[width=.84\linewidth]{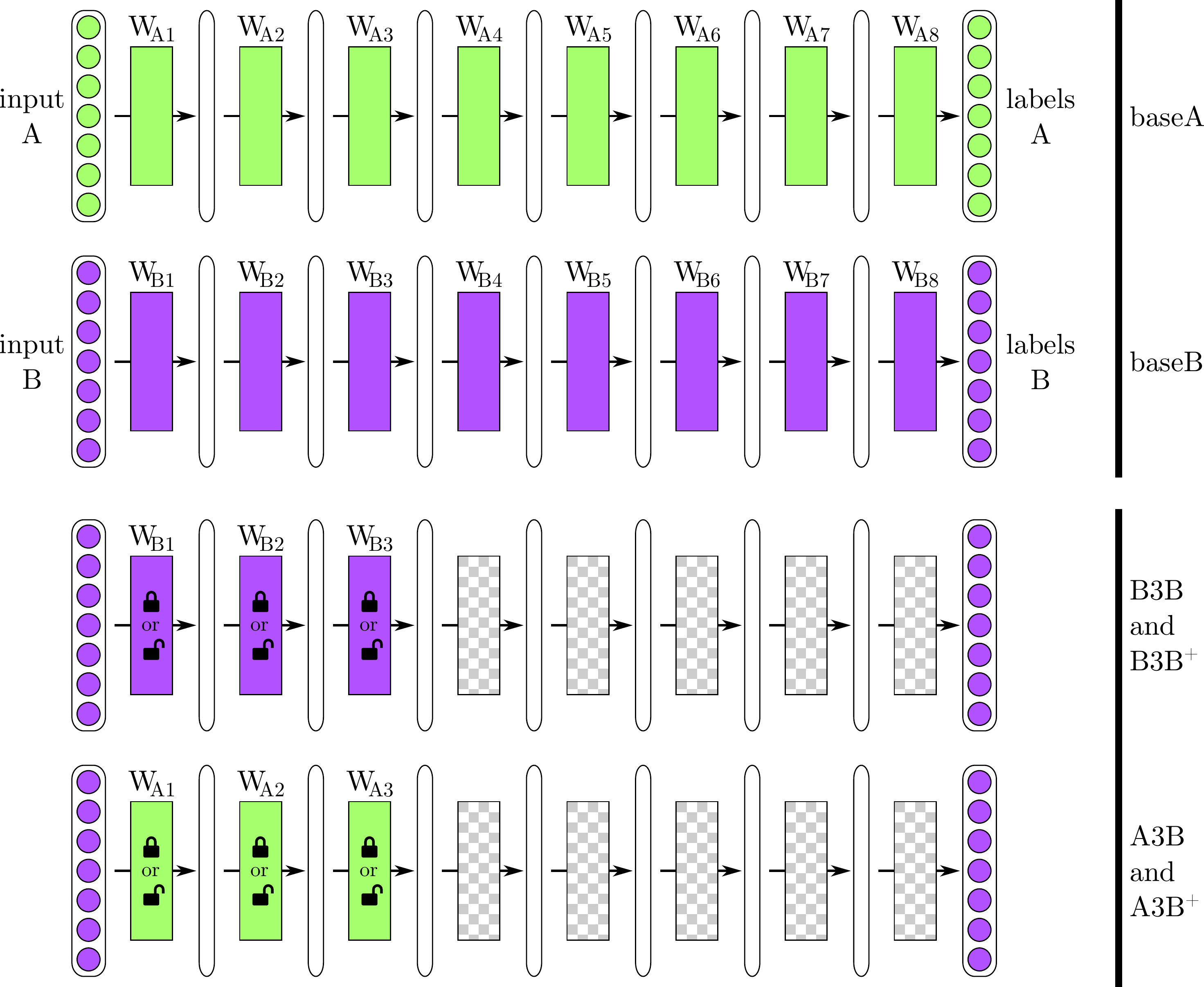}
\end{center}
\caption{Overview of the experimental treatments and controls. \capemph{Top two rows:} The base networks are trained using standard supervised backprop on only half of the ImageNet dataset (first row: \dA half, second row: \dB half).
The labeled rectangles (e.g. $W_{A1}$) represent the weight vector learned for that layer, with the color indicating which dataset the layer was originally trained on. The vertical, ellipsoidal bars between weight vectors represent the activations of the network at each layer.
\capemph{Third row:} In the \emph{selffer} network control, the first $n$ weight layers of the network (in this example, $n=3$) are copied from a base network (e.g. one trained on dataset \dB), the upper $8-n$ layers are randomly initialized, and then the entire network is trained on that same dataset (in this example, dataset \dB). The first $n$ layers are either locked during training (``frozen'' selffer treatment \net{B3B}) or allowed to learn (``fine-tuned'' selffer treatment \net{B3B^+}). This treatment reveals the occurrence of \emph{fragile co-adaptation}, when neurons on neighboring layers co-adapt during training in such a way that cannot be rediscovered when one layer is frozen.
\capemph{Fourth row:} The \emph{transfer} network experimental treatment is the same as the selffer treatment, except that the first $n$ layers are copied from a network trained on one dataset (e.g. \dA) and then the entire network is trained on the \emph{other} dataset (e.g. \dB). This treatment tests the extent to which the features on layer $n$ are general or specific.}
\figlabel{transfer}
\end{figure}

To create base and target datasets that are similar to each other, we randomly assign half of the 1000 ImageNet classes to \dA and half to \dB.
ImageNet contains clusters of similar classes, particularly dogs and cats, like these 13 classes from the biological family \emph{Felidae}:
\{\textit{tabby cat, tiger cat, Persian cat, Siamese cat, Egyptian cat, mountain lion, lynx, 
leopard, snow leopard, jaguar, lion, tiger, cheetah}\}.
On average, \dA and \dB will each contain approximately 6 or 7 of these felid classes, meaning that base networks trained on each dataset will have features at all levels that help classify some types of felids. When generalizing to the other dataset, we would expect that the new high-level felid detectors trained on top of old low-level felid detectors would work well. Thus \dA and \dB are similar when created by randomly assigning classes to each, and we expect that transferred features will perform better than when \dA and \dB are less similar.

Fortunately, in ImageNet we are also provided with a hierarchy of parent classes. This information allowed us to create a special split of the dataset into two halves that are as semantically different from each other as possible: 
with dataset \dA containing only \emph{man-made} entities and \dB containing \emph{natural} entities.
The split is not quite even, with 551 classes in the man-made group and 449 in the natural group. Further details of this split and the classes in each half are given in the supplementary material. In \secref{experiments_nm} we will show that features transfer more poorly (i.e. they are more specific) when the datasets are less similar.

\section{Experimental Setup}
\seclabel{setup}

Since \cite{Krizhevsky-2012} won the ImageNet 2012 competition, there has been much interest and work toward tweaking hyperparameters of large convolutional models. However, in this study 
we aim not to maximize absolute performance, but rather to study transfer results on a well-known architecture.
We use the reference implementation provided by Caffe \citep{jia2014caffe:-convolutional-architecture}
so that our results will be comparable, extensible, and useful to a large number of researchers.
Further details of the training setup (learning rates, etc.) are given in the supplementary material, and code and parameter files to reproduce these experiments are available at \url{http://yosinski.com/transfer}.

\section{Results and Discussion}
\seclabel{experiments}

We performed three sets of experiments. The main experiment has random \dA/\dB splits and is discussed in \secref{experiments_ab}. \secref{experiments_nm} presents an experiment with the man-made/natural split. 
\secref{experiments_random} describes an experiment with random weights.

\subsection{Similar Datasets: Random \dA/\dB splits}
\seclabel{experiments_ab}

\begin{figure}[t]
\begin{center}
\includegraphics[width=1.0\linewidth]{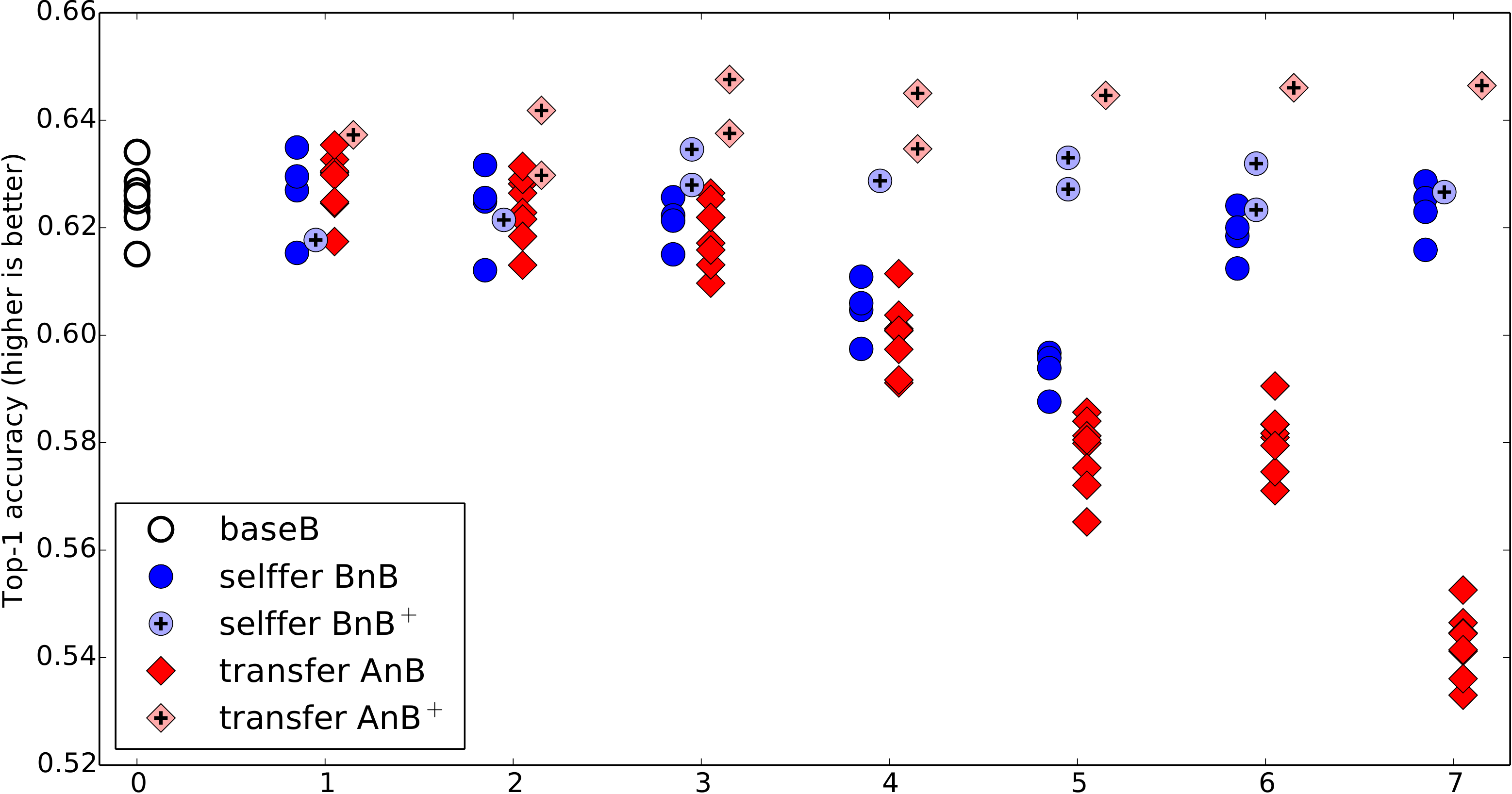}
\includegraphics[width=1.0\linewidth]{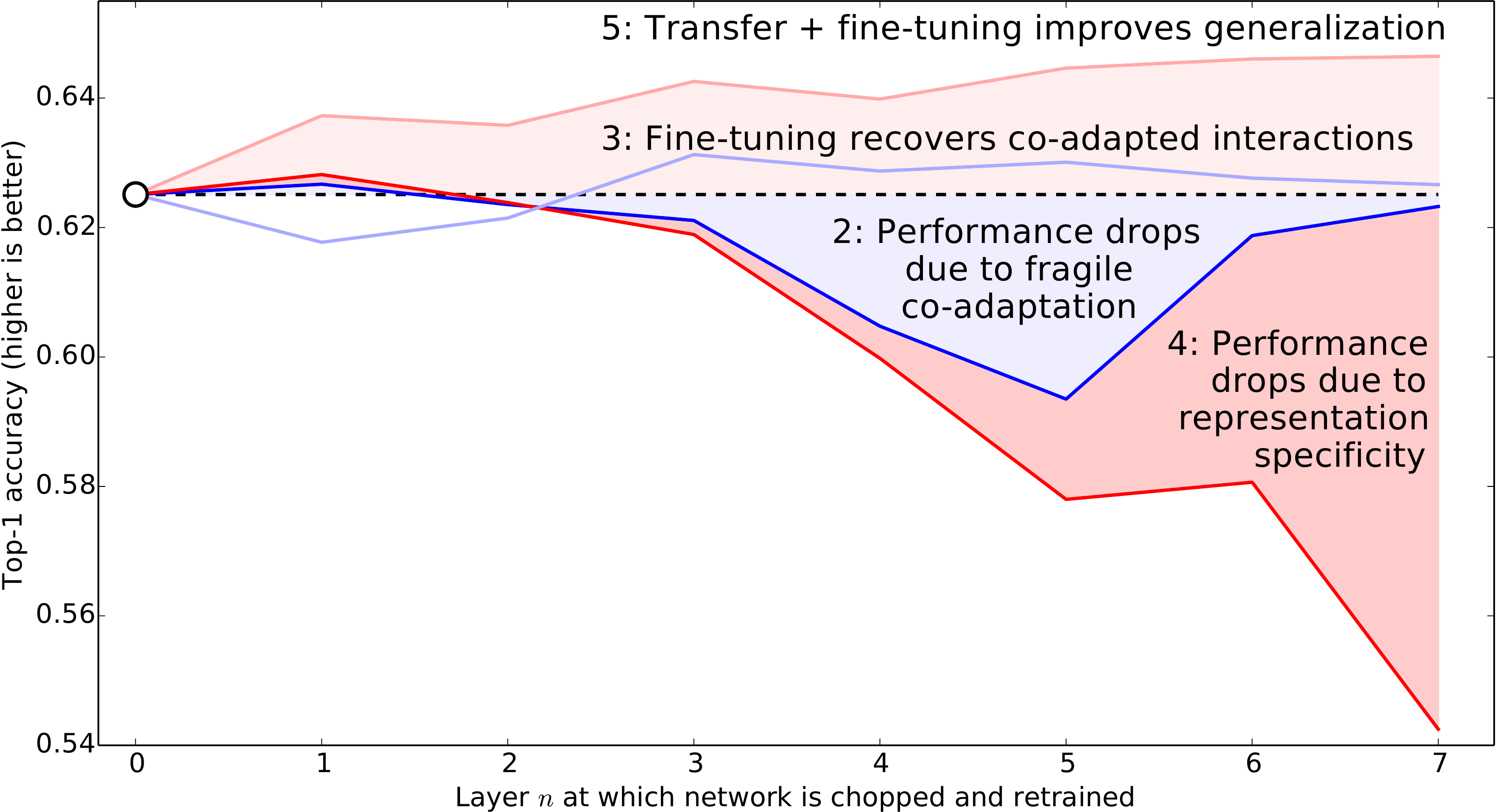}
\end{center}
\caption{The results from this paper's main experiment. \capemph{Top}:
Each marker in the figure represents the average accuracy over the validation set for a trained network. The white circles above $n=0$ represent the accuracy of \net{baseB}. There are eight points, because we tested on four separate random \net{A}/\net{B} splits. Each dark blue dot represents a \net{BnB} network.
Light blue points represent \net{BnB^+} networks, or fine-tuned versions of \net{BnB}. Dark red diamonds are \net{AnB} networks,
and light red diamonds are the fine-tuned \net{AnB^+} versions.
Points are shifted slightly left or right for visual clarity. \capemph{Bottom}: Lines connecting the means of each treatment. Numbered descriptions above each line refer to which interpretation from \secref{experiments_ab} applies.
}
\figlabel{results}
\end{figure}

The results of all \dA/\dB transfer learning experiments on randomly split (i.e. similar) datasets are shown\footnote{\net{AnA} networks and \net{BnB} networks are
 statistically equivalent, because in both cases a network is trained on 500 random classes. To simplify notation we label these \net{BnB} networks. Similarly,
 we have aggregated the statistically identical \net{BnA} and \net{AnB} networks and just call them \net{AnB}.}
 in \figref{results}.
The results yield many different conclusions. In each of the following interpretations, we compare the performance to the base case (white circles and dotted line in \figref{results}).

\begin{enumerate}[leftmargin=1.5em]

\item The white \net{baseB} circles show that a network trained to classify a random subset of 500 classes attains a top-1 accuracy of 0.625, or 37.5\% error. This error is lower than the 42.5\% top-1 error attained on the 1000-class network. While error might have been higher because the network is trained on only half of the data, which could lead to more overfitting, the net result is that error is lower because there are only 500 classes, so there are only half as many ways to make mistakes. 

\item The dark blue \net{BnB} points show a curious behavior. As expected, performance at layer one is the same as the \net{baseB} points. That is, 
if we learn eight layers of features, save the first layer of learned Gabor features and color blobs, reinitialize the whole network, and retrain it toward the same task, it does just as well. This result also holds true for layer 2. However, layers 3, 4, 5, and 6, particularly 4 and 5, exhibit
worse performance.
This performance drop  is evidence that the original network contained \emph{fragile co-adapted features} on successive layers, that is, features that interact with each other in a complex or fragile way such that this co-adaptation \emph{could not be relearned} by the upper layers alone. Gradient descent was able to find a good solution the first time, but this was only possible because the layers were jointly trained. By layer 6 performance is nearly back to the base level, as is layer 7. As we get closer and closer to the final, 500-way softmax output layer 8, there is less to relearn, and apparently relearning these one or two layers is simple enough for gradient descent to find a good solution. Alternately, we may say that there is less co-adaptation of features between layers 6 \& 7 and between 7 \& 8 than between previous layers. To our knowledge it has not been previously observed in the literature that such optimization difficulties may be worse in the middle of a network than near the bottom or top.

\item The light blue \net{BnB^+} points show that when the copied, lower-layer features also learn on the target dataset (which here is the same as the base dataset), performance is similar to the base case. Such fine-tuning thus prevents the performance drop observed in the \net{BnB} networks.

\item The dark red \net{AnB} diamonds show the effect we set out to measure in the first place: the transferability of features from one network to another at each layer. Layers one and two transfer almost perfectly from \net{A} to  \net{B}, giving evidence that, at least for these two tasks, not only are the first-layer Gabor and color blob features general, but the second layer features are general as well. Layer three shows a slight drop, and layers 4-7 show a more significant drop in performance. Thanks to the \net{BnB} points, we can tell that this drop is from a combination of two separate effects: the drop from lost co-adaptation \emph{and} the drop from features that are less and less general. On layers 3, 4, and 5, the first effect dominates, whereas on layers 6 and 7 the first effect diminishes and the specificity of representation dominates the drop in performance.

Although examples of successful feature transfer have been reported elsewhere in the literature \citep{girshick2013rich-feature-hierarchies,donahue2013decaf:-a-deep-convolutional}, to our knowledge these results have been limited to noticing that transfer from a given layer is much better than the alternative of training strictly on the target task, i.e. noticing that the \net{AnB} points at some layer are much better than training all layers from scratch. We believe this is the first time that (1) the extent to which transfer is successful has been carefully quantified layer by layer, and (2) that these two separate effects have been decoupled, showing that each effect dominates in part of the regime.

\item The light red \net{AnB^+} diamonds show a particularly surprising effect: that transferring features and then fine-tuning them results in networks that generalize better than those trained directly on the target dataset. Previously, the reason one might want to transfer learned features is to enable training without overfitting on small target datasets, but this new result suggests that transferring features will boost generalization performance even if the target dataset is large. Note that this effect should not be attributed to the longer total training time (450k base iterations + 450k fine-tuned iterations for \net{AnB^+} vs. 450k for \net{baseB}), because the \net{BnB^+} networks are also trained for the same longer length of time and do not exhibit this same performance improvement. Thus, a plausible explanation is that even after 450k iterations of fine-tuning (beginning with completely random top layers), the effects of having seen the base dataset still linger, boosting generalization performance. It is surprising that this effect lingers through so much retraining. This generalization improvement seems not to depend much on how much of the first network we keep to initialize the second network: keeping anywhere from one to seven layers produces improved performance, with slightly better performance as we keep more layers. The average boost across layers 1 to 7 is 1.6\% over the base case, and the average if we keep at least five layers is 2.1\%.\footnote{We aggregate performance over several layers because each point is computationally expensive to obtain (9.5 days on a GPU), so at the time of publication we have few data points per layer. The aggregation is informative, however, because the performance at each layer is based on different random draws of the upper layer initialization weights. Thus, the fact that layers 5, 6, and 7 result in almost identical performance across random draws suggests that multiple runs at a given layer would result in similar performance.} The degree of performance boost is shown in \tabref{boost}.

\end{enumerate}

%
%
%
%

\begin{table}[t]
\caption{Performance boost of \net{AnB^+} over controls, averaged over different ranges of layers.}
\tablabel{boost}
\begin{center}
\begin{tabular}{|c|c|c|}
\hline
           & mean boost  & mean boost \\
layers     & over        & over \\
aggregated & \net{baseB} & selffer \net{BnB^+} \\
\hline
1-7        & 1.6\%       & 1.4\% \\
3-7        & 1.8\%       & 1.4\% \\
5-7        & 2.1\%       & 1.7\% \\
\hline
\end{tabular}
\end{center}
\end{table}

\subsection{Dissimilar Datasets: Splitting Man-made and Natural Classes Into Separate Datasets }
\seclabel{experiments_nm}

As mentioned previously, the effectiveness of feature transfer is expected to decline as the base and target tasks become less similar.
We test this hypothesis by comparing transfer performance on similar datasets (the random \dA/\dB splits discussed above) to that on dissimilar datasets, created by assigning man-made object classes to \dA and natural object classes to \dB. This man-made/natural split creates datasets as dissimilar as possible within the ImageNet dataset. 

The upper-left subplot of \figref{random_and_nm} shows the accuracy of a \net{baseA} and \net{baseB} network (white circles) and \net{BnA} and \net{AnB} networks (orange hexagons). Lines join common target tasks. The upper of the two lines contains those networks trained toward the target task containing natural categories (\net{baseB} and \net{AnB}). These networks perform better than those trained toward the man-made categories, which may be due to having only 449 classes instead of 551, or simply being an easier task, or both.

\subsection{Random Weights}
\seclabel{experiments_random}

We also compare to random, untrained weights because \cite{Jarrett-ICCV2009} showed --- quite strikingly --- that the combination of random convolutional filters, rectification, pooling, and local normalization can work almost as well as learned features. They reported this result on relatively small networks of two or three learned layers and on the smaller Caltech-101 dataset \citep{Fei-Fei.2004}. It is natural to ask whether or not the nearly optimal performance of random filters they report carries over to a deeper network trained on a larger dataset.

The upper-right subplot of \figref{random_and_nm} shows the accuracy obtained when using random filters for the first $n$ layers for various choices of $n$. Performance falls off quickly in layers 1 and 2, and then drops to near-chance levels for layers 3+, which suggests that getting random weights to work in convolutional neural networks may not be as straightforward as it was for the smaller network size and smaller dataset used by \cite{Jarrett-ICCV2009}. However, the comparison is not straightforward. Whereas our networks have max pooling and local normalization on layers 1 and 2, just as \cite{Jarrett-ICCV2009} did, we use a different nonlinearity ($\mathrm{relu}(x)$ instead of $\mathrm{abs}(\mathrm{tanh}(x))$), different layer sizes and number of layers, as well as other differences. Additionally, their experiment only considered two layers of random weights. The hyperparameter and architectural choices of our network collectively provide one new datapoint, but it may well be possible to tweak layer sizes and random initialization details to enable much better performance for random weights.\footnote{For example, the training loss of the network with three random layers failed to converge, producing only chance-level validation performance. Much better convergence may be possible with different hyperparameters.}

The bottom subplot of \figref{random_and_nm} shows the results of the experiments of the previous two sections after subtracting the performance of their individual base cases. These normalized performances are plotted across the number of layers $n$ that are either random or were trained on a different, base dataset. This comparison makes two things apparent. First, the transferability gap when using frozen features grows more quickly as $n$ increases for dissimilar tasks (hexagons) than similar tasks (diamonds), with a drop by the final layer for similar tasks of only 8\% vs. 25\% for dissimilar tasks. Second, transferring even from a distant task is better than using random filters. One possible reason this latter result may differ from \cite{Jarrett-ICCV2009} is because their fully-trained (non-random) networks were overfitting more on the smaller Caltech-101 dataset than ours on the larger ImageNet dataset, making their random filters perform better by comparison. In the supplementary material, we provide an extra experiment indicating the extent to which our networks are overfit.

\begin{figure}[t]
\begin{center}
\includegraphics[width=1.0\linewidth]{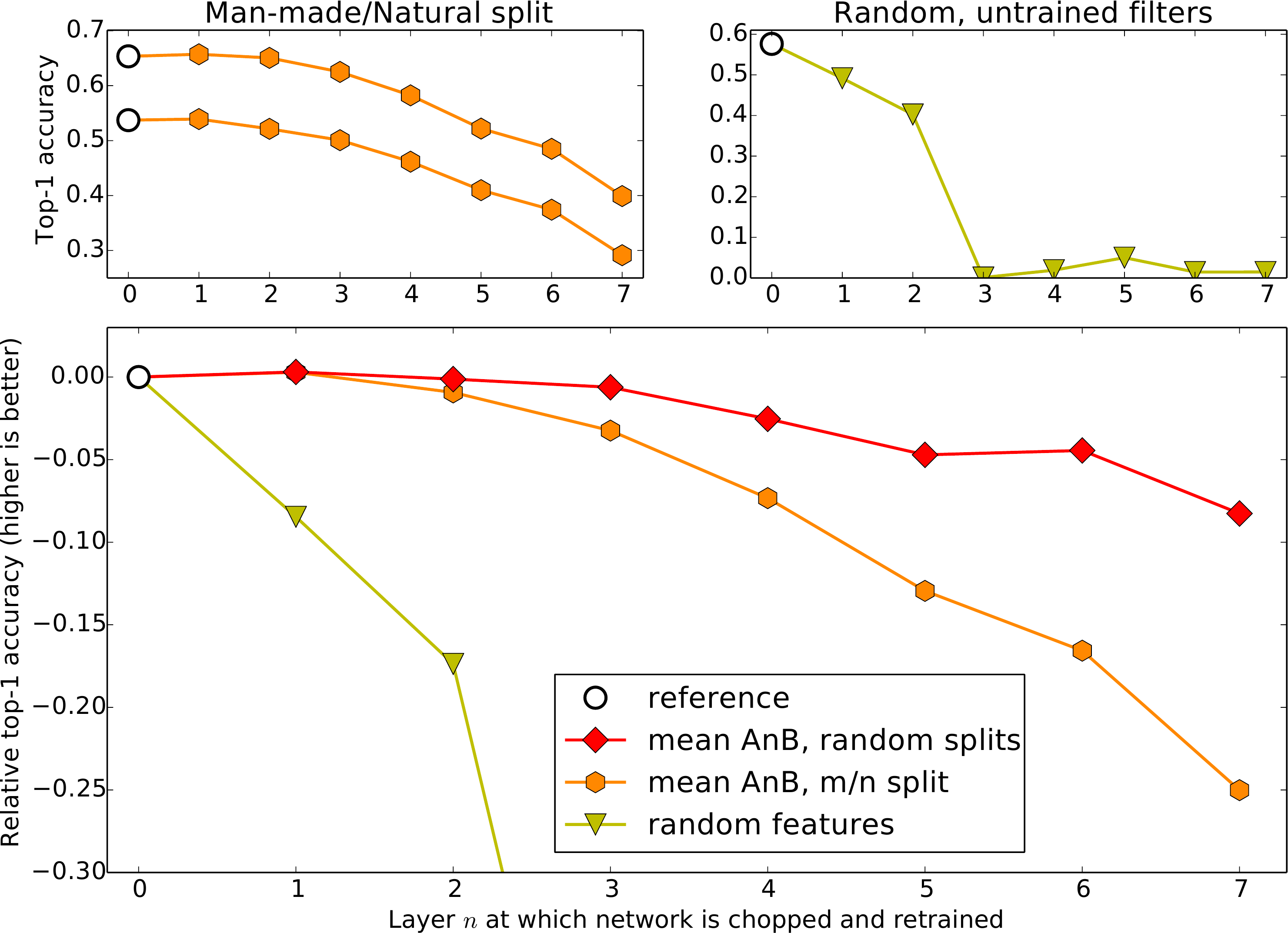}
\end{center}
\caption{Performance degradation vs. layer.
\capemph{Top left}: Degradation when transferring between dissimilar tasks (from man-made classes of ImageNet to natural classes or vice versa). The upper line connects networks trained to the ``natural'' target task, and the lower line connects those trained toward the ``man-made'' target task.
\capemph{Top right}: Performance when the first $n$ layers consist of random, untrained weights.
\capemph{Bottom}: The top two plots compared to the random \dA/\dB split from \secref{experiments_ab} (red diamonds), all normalized by subtracting their base level performance.}
\figlabel{random_and_nm}
\end{figure}

\section{Conclusions}

\vspace*{-1ex}

We have demonstrated a method for quantifying the transferability of features from each layer of a neural network, which reveals their generality or specificity. We showed how transferability is negatively affected by two distinct issues: optimization difficulties related to splitting networks in the middle of fragilely co-adapted layers and the specialization of higher layer features to the original task at the expense of performance on the target task.
We observed that either of these two issues may dominate, depending on whether features are transferred from the bottom, middle, or top of the network.
We also quantified how the transferability gap grows as the distance between tasks increases, particularly when transferring higher layers, but found that even features transferred from distant tasks are better than random weights. Finally, we found that initializing with transferred features can improve generalization performance even after substantial fine-tuning on a new task, which could be a generally useful technique for improving deep neural network performance.


\section*{Acknowledgments}

\vspace*{-1ex}

The authors would like to thank Kyunghyun Cho and Thomas Fuchs for helpful discussions, Joost Huizinga, Anh Nguyen, and Roby Velez for editing, as well as funding from 
the NASA Space Technology Research Fellowship (JY), DARPA project W911NF-12-1-0449, NSERC, Ubisoft, and CIFAR
(YB is a CIFAR Fellow).

{\small
\bibliography{lisa_ml,bibdesk}
\bibliographystyle{natbib}
}

\end{document}


\beginsupplementary

\maketitle



\appendix

\section{Training Details}

Since \cite{Krizhevsky-2012} won the ImageNet 2012 competition, there has naturally been much interest and work toward tweaking hyperparameters of large convolutional models. For example, \cite{Zeiler+et+al-arxiv2013b} found that it is better to decrease the first layer filters sizes from $11\times 11$ to $7\times 7$ and to use a smaller stride of $2$ instead of $4$.
However, because this study aims not for maximum absolute performance but to use 
a commonly studied architecture, we used the reference implementation provided by Caffe \citep{jia2014caffe:-convolutional-architecture}.
We followed \citet{donahue+jia-2013-arxiv} in making a few minor departures from \cite{Krizhevsky-2012} when training the convnets in this study. We skipped the data augmentation trick of adding random multiples of principle components of pixel RGB values, which produced only a $1\%$ improvement in the original paper, and instead of scaling to keep the aspect ratio and then cropping, we warped images to $256\times 256$. We also placed the Local Response Normalization layers just \emph{after} the pooling layers, instead of before them. As in previous studies, including \cite{Krizhevsky-2012}, we use dropout \citep{Hinton-et-al-arxiv2012} on fully connected layers except for the softmax output layer.

We trained with stochastic gradient descent (SGD) with momentum. Each iteration of SGD used a batch size of 256, a momentum of 0.9, and a multiplicative weight decay (for those weights with weight decay enabled, i.e. not for frozen weights) of 0.0005 per iteration.
The master learning rate started at 0.01, and annealed over the course of training by dropping by a factor of 10 every 100,000 iterations. Learning stopped after 450,000 iterations.
Each iteration took about {\raise.17ex\hbox{$\scriptstyle\sim$}}1.7 seconds on a NVidia K20 GPU, meaning the whole training procedure for a single network took {\raise.17ex\hbox{$\scriptstyle\sim$}}9.5 days.


%

Our base model attains a final top-1 error on the validation set of 42.5\%, about the same as the 42.9\% reported by \cite{donahue+jia-2013-arxiv} and 1.8\% worse than \cite{Krizhevsky-2012}, the latter difference probably due to the few minor training differences explained above. We checked these values only to demonstrate that the network was converging reasonably. As our goal is not to improve the state of the art, but to investigate the properties of transfer, small differences in raw performance are not of concern.

Because code is often more clear than text, we've also made all code and parameter files necessary to reproduce these experiments available on
\url{http://yosinski.com/transfer}.

\section{How Much Does an AlexNet Architecture Overfit?}
\seclabel{supp_overfit}

\begin{figure}[th]
\begin{center}
\includegraphics[width=1.0\linewidth]{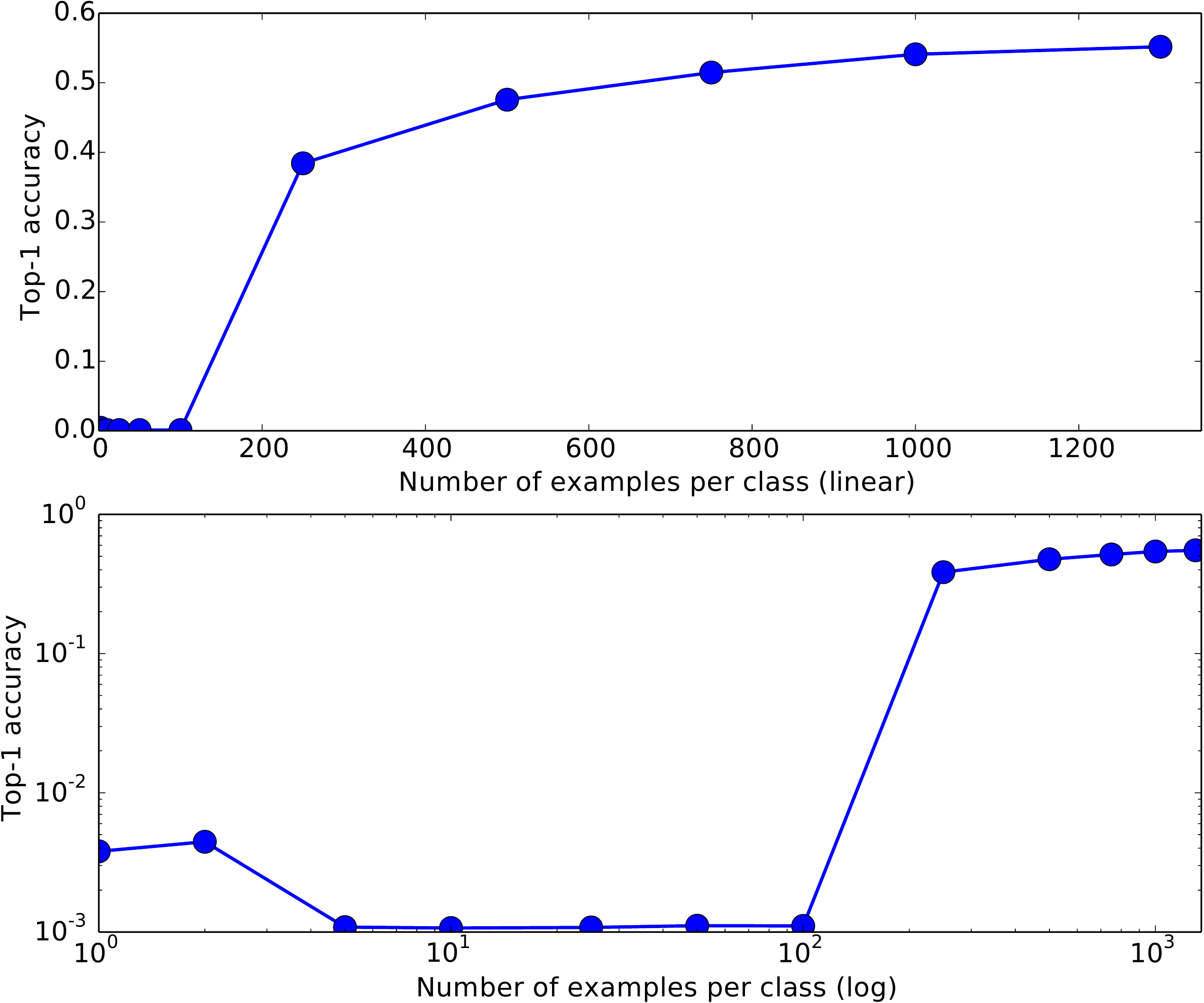}
\end{center}
\caption{Top-1 validation accuracy for networks trained on datasets containing reduced numbers of examples. The largest dataset contains the entire ILSVRC2012 \citep{imagenet_cvpr09} release with a maximum of 1300 examples per class, and the smallest dataset contains only 1 example per class (1000 data points in total). \emph{Top}: linear axes. The slope of the rightmost line segment between 1000 and 1300 is nearly zero, indicating that the amount of overfit is slight. In this region the validation accuracy rises by 0.010820 from 0.54094 to 0.55176. \emph{Bottom}: logarithmic axes. It is interesting to note that even the networks trained on a single example per class or two examples per class manage to attain 3.8\% or 4.4\% accuracy, respectively. Networks trained on \{5,10,25,50,100\} examples per class exhibit poor convergence and attain only chance level performance.}
\figlabel{reduced}
\end{figure}

We observed relatively poor
performance of random filters in an AlexNet
architecture \citep{Krizhevsky-2012} trained on ImageNet, which is in contrast to previously reported successes with random
filters in a smaller convolutional networks
trained on the smaller Caltech-101 dataset \citep{Jarrett-ICCV2009}.
One hypothesis presented in the main paper is that this difference is observed because ImageNet is large
enough to support training an AlexNet architecture without excessive overfitting.
We sought to support or disprove this hypothesis by creating reduced
size datasets containing the same 1000 classes as ImageNet, but where
each class contained a maximum of $n$ examples, for each $n \in \{1300 ,1000 ,750
,500 ,250 ,100 ,50 ,25 ,10 ,5 ,2 ,1\}$. The case of $n = 1300$ is the
complete ImageNet dataset.

Because occupying a whole GPU for this long was infeasible given our available computing resources, we also
devised a set of hyperparameters to allow faster learning by boosting
the learning rate by 25\% to 0.0125, annealing by a factor of 10 after
only 64,000 iterations, and stopping after 200,000 iterations. These
selections were made after looking at the learning curves for the base
case and estimating at which points learning had plateaued and thus annealing could take
place.  This faster training schedule was only used for the
experiments in this section. Each run took just over 4 days on a K20
GPU.

\begin{table}[th]
\caption{An enumeration of the points in \figref{reduced} for clarity.}
\tablabel{reduced}
\begin{center}
\begin{tabular}{|r|l|}
\hline
Number        & Top-1 \\
 of examples  & validation \\
per class           & accuracy \\
\hline
1300  &  0.55176  \\
1000  &  0.54094  \\
750   &  0.51470   \\
500   &  0.47568   \\
250   &  0.38428   \\
100   &  0.00110   \\
50    &  0.00111   \\
25    &  0.00107   \\
10    &  0.00106   \\
5     &  0.00108   \\
2     &  0.00444   \\
1     &  0.00379   \\
\hline
\end{tabular}
\end{center}
\end{table}

The results of this experiment are shown in \figref{reduced}
and \tabref{reduced}. The rightmost few points in the top subplot
of \figref{reduced} appear to converge, or nearly converge, to an
asymptote, suggesting that validation accuracy would not improve
significantly when using an AlexNet model with much more data, and
thus, that the degree of overfit is not severe.

\section{Man-made vs. Natural Split}
\seclabel{supp_natman}

In order to compare transfer performance between tasks \dA and \dB
such that \dA and \dB are as semantically dissimilar as possible, we
sought to find two disjoint subsets of the 1000 classes in ImageNet
that were as unrelated as possible. To this end we annotated each node
$x_i$ in the WordNet graph with a label $n_i$ such that $n_i$ is the
number of distinct ImageNet classes reachable by starting at $x_i$ and traversing the graph only in the parent
$\rightarrow$ child direction.
The 20 nodes with largest $n_i$ are the following:

{
\small
\begin{verbatim}
 n_i   x_i
1000   n00001740: entity
 997   n00001930: physical entity
 958   n00002684: object, physical object
 949   n00003553: whole, unit
 522   n00021939: artifact, artefact
 410   n00004475: organism, being
 410   n00004258: living thing, animate thing
 398   n00015388: animal, animate being, beast, brute, creature, fauna
 358   n03575240: instrumentality, instrumentation
 337   n01471682: vertebrate, craniate
 337   n01466257: chordate
 218   n01861778: mammal, mammalian
 212   n01886756: placental, placental mammal, eutherian, eutherian mammal
 158   n02075296: carnivore
 130   n03183080: device
 130   n02083346: canine, canid
 123   n01317541: domestic animal, domesticated animal
 118   n02084071: dog, domestic dog, Canis familiaris
 100   n03094503: container
  90   n03122748: covering
\end{verbatim}
}

Starting from the top, we can see that the largest subset, \texttt{entity}, contains all
1000 ImageNet categories. Moving down several items, the first
subset we encounter containing approximately half of the classes
is \texttt{artifact} with 522 classes. The next is \texttt{organism}
with 410. Fortunately for this study, it just so happens that these two subsets are mutually
exclusive, so we used the first to populate our \textit{man-made} category and the second to populate our \textit{natural} category.
 There are
$1000 - 522 - 410 = 68$ classes remaining outside these two subsets, and we manually assigned these to either category as seemed more appropriate.
For example, we placed
\texttt{pizza}, 
\texttt{cup}, and
\texttt{bagel} into \textit{man-made} and 
\texttt{strawberry}, 
\texttt{volcano}, and 
\texttt{banana} into \textit{natural}. This process results in 551 and 449 classes, respectively. The 68 manual decisions are shown below, and the complete list of 551 man-made and 449 natural classes is available at \url{http://yosinski.com/transfer}.

\pagebreak

Classes manually placed into the man-made category:
{
\scriptsize
\begin{verbatim}
n07697537 hotdog, hot dog, red hot
n07860988 dough
n07875152 potpie
n07583066 guacamole
n07892512 red wine
n07614500 ice cream, icecream
n09229709 bubble
n07831146 carbonara
n07565083 menu
n07871810 meat loaf, meatloaf
n07693725 bagel, beigel
n07920052 espresso
n07590611 hot pot, hotpot
n07873807 pizza, pizza pie
n07579787 plate
n06874185 traffic light, traffic signal, stoplight
n07836838 chocolate sauce, chocolate syrup
n15075141 toilet tissue, toilet paper, bathroom tissue
n07613480 trifle
n07880968 burrito
n06794110 street sign
n07711569 mashed potato
n07932039 eggnog
n07695742 pretzel
n07684084 French loaf
n07697313 cheeseburger
n07615774 ice lolly, lolly, lollipop, popsicle
n07584110 consomme
n07930864 cup
\end{verbatim}
}

Classes manually placed into the natural category:

{
\scriptsize
\begin{verbatim}
n13133613 ear, spike, capitulum
n07745940 strawberry
n07714571 head cabbage
n09428293 seashore, coast, seacoast, sea-coast
n07753113 fig
n07753275 pineapple, ananas
n07730033 cardoon
n07749582 lemon
n07742313 Granny Smith
n12768682 buckeye, horse chestnut, conker
n07734744 mushroom
n09246464 cliff, drop, drop-off
n11879895 rapeseed
n07718472 cucumber, cuke
n09468604 valley, vale
n07802026 hay
n09288635 geyser
n07720875 bell pepper
n07760859 custard apple
n07716358 zucchini, courgette
n09332890 lakeside, lakeshore
n09193705 alp
n09399592 promontory, headland, head, foreland
n07717410 acorn squash
n07717556 butternut squash
n07714990 broccoli
n09256479 coral reef
n09472597 volcano
n07747607 orange
n07716906 spaghetti squash
n12620546 hip, rose hip, rosehip
n07768694 pomegranate
n12267677 acorn
n12144580 corn
n07718747 artichoke, globe artichoke
n07753592 banana
n09421951 sandbar, sand bar
n07715103 cauliflower
n07754684 jackfruit, jak, jack
\end{verbatim}
}

\pagebreak

\bibliography{lisa_ml,bibdesk}
\bibliographystyle{natbib}